\documentclass[runningheads]{llncs}
\usepackage[T1]{fontenc}
\usepackage{graphicx}
\usepackage{amsmath,amssymb}
\usepackage{booktabs}
\usepackage{xspace}
\usepackage{tikz}
\usetikzlibrary{arrows.meta,positioning,fit,backgrounds,calc}
\usepackage[hidelinks]{hyperref}  
\usepackage{orcidlink}
\usepackage{bbding}              
\usepackage{makecell}
\usepackage{multirow}
\usepackage{placeins}

\usepackage{xcolor}

\newcommand{\FMMF}{0.874\xspace}
\newcommand{\FMBA}{0.860\xspace}  
\newcommand{\FMGV}{0.784\xspace} 
\newcommand{\dragClass}{Gleason\,5\xspace} 
\newcommand{\MF}{0.894\xspace}
\newcommand{\GV}{0.802\xspace}
\newcommand{\BA}{0.926\xspace}

\newcommand{\B}[1]{\textcolor{red}{\textbf{#1}}}

\begin{document}

\title{Multi-scale Fusion of Fast Vision Mamba and Foundation Model Features for Patch-Level Gleason Grading in Computational Pathology}

\author{%
Zihe Zhang\inst{1}\orcidlink{0009-0002-0819-6522} \and
Weimiao Yu\inst{2}\orcidlink{0000-0001-5092-1009} \and
Piumi Don Simonge\inst{2}\orcidlink{0000-0003-0877-1128}\textsuperscript{(\Envelope)}}
\authorrunning{Z. Zhang et al.}
\titlerunning{Multi-scale Fusion of FastVim for Patch-Level Gleason Grading}
\institute{%
Nanyang Technological University, Singapore\\
\email{ZH0013HE@e.ntu.edu.sg}
\and
Bioinformatics institute, Agency for Science Technology and Research, Singapore\\
\email{yu\_weimiao@a-star.edu.sg}, \email{piumi\_simonge@a-star.edu.sg}}
\maketitle

\begin{abstract}
Automated Gleason grading of prostate whole slide images (WSIs) enables faster and more consistent cancer risk stratification. However, it remains a challenging task due to the subtle morphological differences between grades, the severe class imbalance arising from rare patterns, and the diagnostic features that span multiple spatial scales. Pathology foundation models (FMs) extract strong patch representations but often operate at a single-scale, which limits spatial context and disproportionately affects classes that depend on broader context. The challenge of fusing multiple scales and complementary models, with minimal data and computational resources, remains unresolved. We propose a multi-scale classifier built on state-space models. Our method combines two complementary feature sources: a frozen foundation model, and a lightweight FastVim, pre-trained on ImageNet and distilled from the foundation model, that reads the raw image crops. At each scale, a Mamba model encodes the features from both sources with linear-time spatial context, and a jointly trained head fuses the encoded representations into one prediction.
On the PANDA dataset, our method reaches macro F1 of \MF and \dragClass F1 of \GV under 5-fold cross-validation, a 12\% and 21\% relative improvement over the best single-scale model. A controlled study over 5 foundation models, 5 base classifiers and 3 scales confirms that the gain holds across encoders, with larger improvements observed at smaller scales.
\end{abstract}

\section{Introduction}
Prostate cancer is the second most commonly diagnosed cancer in men worldwide~\cite{sung2021globocan}. Gleason grading of biopsy tissue provides essential prognostic information that guides treatment decisions~\cite{epstein2016isup}, but it is labour intensive and shows considerable inter-observer variability~\cite{vanderlaak2021survey}. Gleason grading assigns prostate cancer tissue into 3 main architectural patterns: Gleason pattern 3, 4, and 5, where higher patterns correspond to progressively poorer glandular differentiation, accompanied by more pronounced cellular atypia~\cite{epstein2016isup}. Reliable and reproducible automation of this assessment is therefore clinically valuable.

Deep learning is the standard tool for histopathology image analysis~\cite{vanderlaak2021survey}, and pathology foundation models (FMs) have become its dominant approach. Pre-trained with heavy self-supervision on large datasets, FMs produce rich patch representations that transfer across downstream tasks. UNI~\cite{chen2024uni} pre-trains a ViT-L/16 with DINOv2~\cite{oquab2023dinov2} on over 100{,}000 whole slide images (WSIs), and its successor UNI2 scales further. Virchow~\cite{vorontsov2024virchow} scales to a 632M parameter ViT-H/14 trained on 1.5M slides, Virchow2~\cite{zimmermann2024virchow2} adds mixed magnification training, and CONCH~\cite{lu2024conch} adds vision-language pre-training.

State-of-the-art FMs are trained at a pre-determined patch size and lose accuracy at substantially different scales~\cite{mollers2026mindthegap}. However, Gleason grading is intrinsically multi-scale, relying on cues from cellular detail to glandular and tissue architecture. Since WSIs contain hundreds of thousands of patches, modelling spatial context with attention is quadratic and quickly becomes computationally prohibitive. State-space models (SSMs) such as Mamba~\cite{gu2023mamba} process long token sequences in linear time. The vision counterpart of Mamba, Vision Mamba (ViM)~\cite{zhu2024vim}, extends this architecture to natural image understanding, while pathology-specific variant Vim4Path~\cite{nasirisarvi2024vim4path} adapts this architecture for computational pathology. FastVim~\cite{kapse2025fastvim} improves on ViM, maintaining accuracy while achieving up to $72.5\%$ faster inference even at high resolution ($2048\times2048$). Beyond efficient context modelling, the remaining question is how to combine scales. Prior multi-scale analysis examines several patch sizes but typically combines the single-scale predictions by ensembling or averaging~\cite{das2017fusion,koziarski2024diagset}, which treats the scales independently and discards their joint structure. Multi-scale fusion of histopathology images in an end-to-end model still remains an open challenge.

To close this gap, we formulate multi-scale patch classification as a single, jointly trained network. Our method draws features from two complementary sources: a frozen FM, and a lightweight FastVim, pre-trained on ImageNet and distilled from the FM, that reads the raw image crops the pooled FM features miss. At each scale, a Mamba encoder adds spatial context, and a jointly trained head fuses the representations from both sources into one prediction, rather than averaging or voting on single-scale outputs. Full details are given in Sec.~\ref{sec:method}.

We validate our method on the PANDA dataset~\cite{bulten2022panda}, a large-scale public prostate cohort with pixel-level masks, on a 5-class tissue classification task (see Sec.~\ref{sec:exp:data}). Under 5-fold cross-validation, our method increases macro F1 from $0.796$ for the best single-scale model to \MF, a $12\%$ relative gain, and Gleason 5 F1 from $0.661$ to \GV, a $21\%$ relative gain (Table~\ref{tab:main}). With minimal resources, our FastVim branch reaches higher throughput than a same-size vision transformer at these input sizes (Table~\ref{tab:backbone}), and it is trained on the target cohort alone.

\begin{figure}[t]\centering\includegraphics[width=\textwidth]{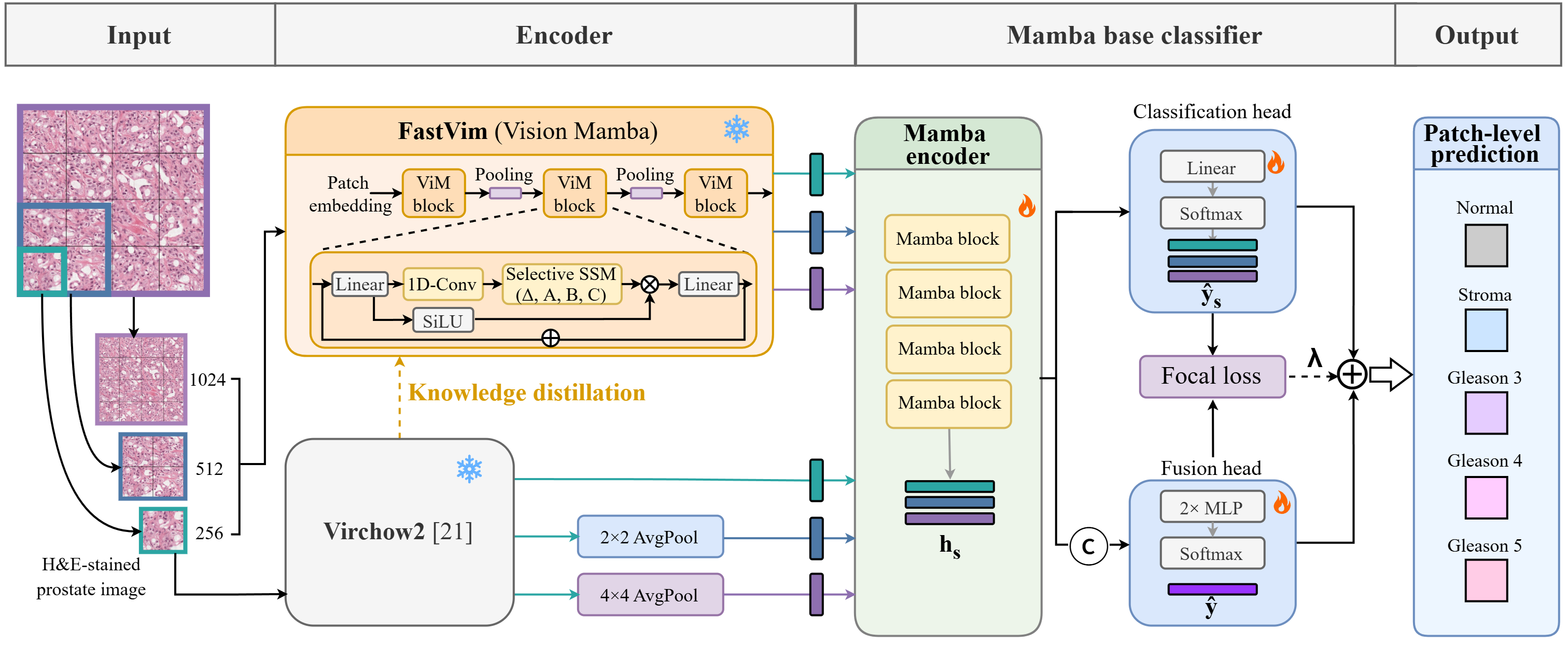}
\caption {Overview of the proposed method. Since Virchow2 accepts only its native $256$\,px input, the $512$\,px and $1024$\,px scale representations are formed by averaging the embeddings of the $2\times2$ and $4\times4$ constituent tiles, respectively. In contrast, the FastVim branch encodes native crops at all three scales directly. The representations from both branches are each encoded by a Mamba encoder and fused into one patch-level prediction, and a per-scale classification head supplies an auxiliary training signal (deep supervision). Only the Mamba encoders, the per-scale classification heads and the fusion head are trained, and the two feature encoders are kept frozen.}
\label{fig:overview}
\end{figure}

\section{Method}
\label{sec:method}

 \subsection{Overview}
We frame patch classification as a multi-scale problem, describing each patch at three spatial scales. Our method (Fig.~\ref{fig:overview}) has three parts. Two frozen encoders produce patch features at every scale: a pretrained foundation model, and a lightweight FastVim distilled from it that reads the raw image crops (Sec.~\ref{sec:method:feat} and~\ref{sec:method:branch}). A Mamba encoder at each scale then adds spatial context from neighbouring patches (Sec.~\ref{sec:method:fuse}). Finally, a fusion head combines the encoded representations into a patch-level prediction, and a per-scale classification head provides deep supervision (Sec.~\ref{sec:method:train}).

\subsection{Multi-scale Feature Extraction}
\label{sec:method:feat}
The FM branch uses Virchow2~\cite{zimmermann2024virchow2}, which achieved best performance among 5 state-of-the-art FMs (more details in Sec.~\ref{sec:exp:branch}). We feed it patches at its recommended input size of $256$\,px ($0.48\,\mu$m/pixel, $20\times$) and each patch is converted to a $1280$-dimensional embedding. The larger scales ($512$ and $1024$\,px) are defined as the mean of these $256$\,px embeddings over non-overlapping $2\times2$ and $4\times4$ grids. Let $S$ denote the number of scales, with $S{=}3$ throughout. Together, the scales capture local cellular detail, glandular structure, and broader tissue architecture.

\subsection{FastVim Branch}
\label{sec:method:branch}
The pooling above discards fine morphological detail present in the original crops, which cannot be recovered downstream. We therefore add a second branch that reads the image crops directly. To keep it computationally feasible, we use a lightweight ViM, FastVim~\cite{kapse2025fastvim}, pre-trained on ImageNet. Among the backbones we compared, FastVim offers the best trade-off between accuracy and throughput at larger inputs (Sec.~\ref{sec:exp:branch}, Table~\ref{tab:backbone}). It is trained on crops at $S$ scales, each fed at its native size except the smallest patch size ($256$\,px), which is resized to $224$\,px.

We distill Virchow2 into this branch via knowledge distillation~\cite{hinton2015distilling}, applied independently at each scale. The teacher is the frozen FM with an MLP classifier trained on the same patches, which gives soft targets $\mathbf{q}^{(s)}=\mathrm{softmax}(\boldsymbol{\ell}^{(s)}/T)$ at each scale $s$ and temperature $T$, with $T>1$ so the targets reflect relative class probabilities rather than a hard label. The student, with logits $\hat{\boldsymbol{\ell}}^{(s)}$, is trained from its ImageNet initialisation with the loss $\mathcal{L}_{\mathrm{distill}}$. This is calculated as the weighted sum of a cross-entropy term (CE) and a Kullback-Leibler divergence term (KL), with weight $\alpha$ and ground truth label $\mathbf{y}$.
\begin{equation}
\mathcal{L}_{\mathrm{distill}} = \alpha\,\mathrm{CE}\!\left(\mathbf{y}, \mathrm{softmax}(\hat{\boldsymbol{\ell}}^{(s)})\right) + (1-\alpha)\, T^2\, \mathrm{KL}\left(\mathbf{q}^{(s)} \,\middle\|\, \mathrm{softmax}(\hat{\boldsymbol{\ell}}^{(s)}/T)\right)
\end{equation}
Once trained, the student is frozen and never updated again. We take its penultimate pooled features, of dimension $d_{\mathrm{ViM}}{=}384$ against $d_{\mathrm{FM}}{=}1280$ for the FM branch. Both branches then stay frozen for the remainder of training (Sec.~\ref{sec:method:fuse}).

\subsection{Scale Encoders and Fusion Head}
\label{sec:method:fuse}
Each branch $b\in\mathcal{B}=\{\mathrm{FM},\mathrm{ViM}\}$ provides a $d_{b}$-dimensional feature for every patch at each scale $s$. We tile each slide into fixed-length sequences of $N{=}256$ patches in row-major order, where $N$ is unrelated to the patch size, and repeat the coarser-scale features across the finest-scale patches they cover so that all sequences are aligned on the same patch grid.

Each branch $b$ and scale $s$ has its own Mamba encoder~\cite{gu2023mamba} $f_{b}^{(s)}:\mathbb{R}^{N\times d_{b}}\!\to\!\mathbb{R}^{N\times d'}$, with $d'{=}256$, which adds spatial context from neighbouring patches in $O(N)$ time and memory. Weights are not shared across scales or across branches, since the two branches differ in feature dimension. We write $\mathbf{h}_{b}^{(s)}\in\mathbb{R}^{d'}$ for the row of the output that corresponds to a given patch.

These representations are concatenated into $\mathbf{z}$, where
\begin{equation}
\mathbf{z} = \big[\,
\mathbf{h}_{\mathrm{FM}}^{(1)} \Vert \cdots \Vert \mathbf{h}_{\mathrm{FM}}^{(S)}
\Vert
\mathbf{h}_{\mathrm{ViM}}^{(1)} \Vert \cdots \Vert \mathbf{h}_{\mathrm{ViM}}^{(S)}
\,\big] \in \mathbb{R}^{|\mathcal{B}|Sd'}
\end{equation}
and a two-layer fusion head $g:\mathbb{R}^{|\mathcal{B}|Sd'}\!\to\!\Delta^{C-1}$ maps $\mathbf{z}$ to a distribution over the $C{=}5$ classes. A per-scale classification head predicts from $[\mathbf{h}_{\mathrm{FM}}^{(s)}\,\|\,\mathbf{h}_{\mathrm{ViM}}^{(s)}]$ alone, giving deep supervision during training (Sec.~\ref{sec:method:train}).

Because the fusion head combines the full representations before any single-scale decision, it is more expressive than a fixed scale. A uniform average of predictions weights every class and scale equally, whereas the fusion head learns weights that depend on the class and the scale. Sec.~\ref{sec:exp:main} reports its margin over this uniform average baseline. The fusion head adds under $1\%$ of the parameters and inference time of the Mamba encoders.

\subsection{Training and Computational Cost}
\label{sec:method:train}
We first train FastVim on image crops with distillation from the frozen FM, using temperature $T{=}4$ and distillation weight $\alpha{=}0.5$. We then train the Mamba encoders, the classification heads and the fusion head with the loss $\mathcal{L}$, computed as the sum of a focal loss on the fused prediction and focal losses on the individual predictions at each scale,
\begin{equation}
\mathcal{L} = \mathcal{L}_{\text{focal}}(\hat{y}) + \lambda \sum_{s} \mathcal{L}_{\text{focal}}\big(\hat{y}^{(s)}\big),
\end{equation}
where $\hat{y}$ is the fused prediction, $\hat{y}^{(s)}$ is the prediction of the classification head at scale $s$, and $\lambda{=}0.3$. Both terms are focal losses ($\gamma{=}1$) with class weights $w_k\propto 1/\sqrt{n_k}$, where $n_k$ is the number of patches of class $k$, to counter the class imbalance. We optimise with AdamW~\cite{loshchilov2019decoupled} at learning rate $10^{-4}$, weight decay $10^{-4}$, batch size $32$, for $50$ epochs with cosine scheduling, and keep the checkpoint with the best validation balanced accuracy. Each Mamba encoder has four layers with hidden dimension $256$ and state size $16$, and the fusion head is a two-layer MLP with hidden dimension $128$.

\section{Experiments and Results}
\label{sec:exp}
\subsection{Dataset and Evaluation}
\label{sec:exp:data}
Unless otherwise stated, all experiments use Virchow2, the strongest of the 5 FMs we compare (Sec.~\ref{sec:exp:analysis}). We evaluate our method on the PANDA dataset~\cite{bulten2022panda}, which provides $10{,}508$ slides, of which $10{,}497$ have pixel-level tissue masks. We hold out a fixed test set of $2{,}100$ slides and split the remaining $8{,}397$ into 5 folds. Each fold trains on 4 parts and validates on the fifth, giving 5 models that we evaluate on the same held-out test set and average. Splitting is stratified at the slide level by each slide's major Gleason grade, preserving the class distribution. From each slide we extract patches at the $256$, $512$, and $1024$ px scales (Sec.~\ref{sec:method:feat}) and label each patch by the majority class above a $70\%$ threshold, giving about $3.8$M patches at the $256$ scale. The dataset is severely imbalanced, with $79\%$ normal, $16\%$ stroma, $1.8\%$ Gleason 3 (G3), $3.4\%$ Gleason 4 (G4), and $0.5\%$ Gleason 5 (G5). Fig.~\ref{fig:splits} in the supplementary material shows the data splits across main experiment and ablation studies (Sec.~\ref{sec:exp:branch}) as well as the class distribution across each split. We report balanced accuracy, macro F1, and per-class F1. F1 of the minority class (Gleason 5) is a primary criterion for assessing performance under class imbalance.

\subsection{Multi-scale Fusion Results}
\label{sec:exp:main}
\begin{figure}[t]\centering
\includegraphics[width=0.5\textwidth]{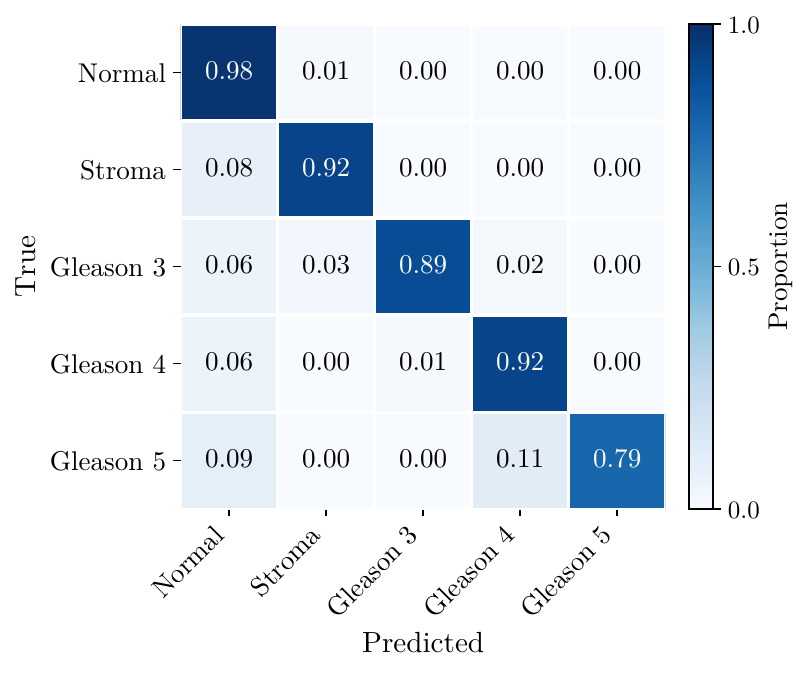}
\caption{Confusion matrix showing that our method classifies most patches correctly and that the remaining errors fall between adjacent Gleason patterns. Each cell is the proportion of a true class (row) predicted as each class (column), so the diagonal gives per-class recall.}
\label{fig:confusion}
\end{figure}

Table~\ref{tab:main} compares the performance of our method with single-scale baselines and a multi-scale uniform average, which averages the class probabilities of the three single-scale models and weights every scale and class equally. The best single-scale reaches macro F1 of $0.796$ and Gleason 5 F1 of $0.661$, and the uniform average does not improve on this (macro F1 = $0.747$), being dominated by the majority-heavy coarse scale. Our method increases macro F1 to \FMMF and Gleason 5 F1 to \FMGV, with consistent improvements ($\pm0.001$ and $\pm0.005$) across all folds. However, this improvement is accompanied by a marginal reduction of majority-class recall, resulting in a slightly reduced balanced accuracy of \FMBA. By adding the FastVim branch, balanced accuracy is restored to \BA, on par with the best single-scale, while macro F1 and Gleason 5 F1 increase further to \MF and \GV. The confusion matrix (Fig.~\ref{fig:confusion}) shows the remaining errors fall mostly between adjacent Gleason patterns. Further visual inspection shows that some high-grade patches predicted as normal lie at tumour boundaries and appear to be affected by annotation noise, suggesting that part of the balanced accuracy gap reflects label noise rather than model error.

\begin{table}[t]\centering
\caption{Performance of our multi-scale method compared with single-scale baselines and a multi-scale uniform average. All experiments use the best-performing foundation model, Virchow2, where applicable. Results are mean $\pm$ standard deviation over the 5 folds on the held-out test set. Best value per column in bold.}
\label{tab:main}
\setlength{\tabcolsep}{4pt}
\begin{tabular}{lccc}
\toprule
Model & Macro F1 & F1 (G5) & Bal. Acc \\
\midrule
Single-scale, $256$              & $0.794{\pm}0.012$ & $0.636{\pm}0.044$ & $0.916{\pm}0.003$ \\
Single-scale, $512$              & $0.796{\pm}0.016$ & $0.661{\pm}0.036$ & $0.922{\pm}0.004$ \\
Single-scale, $1024$             & $0.768{\pm}0.028$ & $0.626{\pm}0.080$ & $\mathbf{0.929}{\pm}0.003$ \\
Multi-scale (uniform average)    & $0.747{\pm}0.003$ & $0.535{\pm}0.019$ & $0.923{\pm}0.001$ \\
\specialrule{0.02em}{0pt}{1pt}
Multi-scale (ours)               & $0.874{\pm}0.001$ & $0.784{\pm}0.005$ & $0.860{\pm}0.005$ \\
Multi-scale + FastVim (ours)           & $\mathbf{0.894}{\pm}0.002$ & $\mathbf{0.802}{\pm}0.003$ & $0.926{\pm}0.004$ \\
\bottomrule
\end{tabular}
\end{table}

\subsection{Ablation Studies}
\label{sec:exp:branch}
This subsection evaluates two design choices in the absence of the FM branch, comparing candidate backbones for the FastVim branch (Table~\ref{tab:backbone}) and testing whether multi-scale fusion generalises across different image encoders (Table~\ref{tab:agnostic}). To limit the computational cost, the ablation experiments use a single stratified train/validation/test split ($70/15/15$), which ensures the patch-level class proportion is within $0.54$ percentage points of the whole cohort. Since the ablation experiments use a different dataset split from Sec.~\ref{sec:exp:main}, absolute values are not comparable across the two protocols. 

Table~\ref{tab:backbone} compares candidate backbones for the FastVim branch at single-scale ($1024$ crops). The candidates cover the vision transformer, state-space and efficient state-space models at $22 - 26$M parameters, so the comparison isolates the architecture rather than the capacity. Since the Gleason patterns follow a graded histological severity, we report quadratic weighted kappa (QWK), which penalises misclassifications in proportion to their distance from the true grade. We also report throughput, as this branch reads the crops natively and its cost grows with scale. FastVim-S reaches the highest macro F1 and QWK at about $1.5\times$ the throughput of ViM-S. Distilling Virchow2 into FastVim-S increases its macro F1 from $0.665$ to $0.700$ and QWK from $0.631$ to $0.672$.

Table~\ref{tab:agnostic} shows that applying the proposed multi-scale fusion to four image encoders consistently improves both macro F1 and Gleason 5 F1. Each encoder is frozen and provides features on the $256$\,px patches. In Table~\ref{tab:agnostic} single-scale column uses these features alone, and the fusion column adds the pooled $512$ and $1024$\,px scales. All downstream components remain unchanged. For Virchow2, macro F1 and Gleason 5 F1 increase from $0.717$ to $0.852$ and from $0.514$ to $0.730$, respectively. Similar improvements are observed across the vision
transformer~\cite{touvron2021training}, the convolutional neural
network~\cite{he2016deep}, and the state-space~\cite{kapse2025fastvim}
encoder. Therefore, we conclude that the fusion improves minority-class performance regardless of the underlying encoder.

\begin{table}[t]\centering
\caption{Backbones for the FastVim branch at $1024$ crops ($70/15/15$ split, not comparable with Table~\ref{tab:main}). Throughput is measured in patches per second on an A100 GPU.}
\label{tab:backbone}
\setlength{\tabcolsep}{6pt}
\begin{tabular}{lcccc}
\toprule
Backbone & Macro F1 & QWK & Throughput & \#Params \\
\midrule
DeiT-S (ViT)~\cite{touvron2021training}      & 0.621 & 0.597 & 47 & 22M \\
ViM-S~\cite{zhu2024vim}             & 0.625 & 0.598 & 43 & 26M \\
FastVim-S~\cite{kapse2025fastvim}         & 0.665 & 0.631 & 63 & 26M \\
FastVim-S with distillation & \textbf{0.700} & \textbf{0.672} & 63 & 26M \\
\bottomrule
\end{tabular}
\end{table}

\begin{table}[t]\centering
\caption{Multi-scale fusion applied to different image encoders, on the $70/15/15$ split. Best results per metric in bold.}
\label{tab:agnostic}
\resizebox{\textwidth}{!}{%
\setlength{\tabcolsep}{5pt}
\begin{tabular}{lcccccc}
\toprule
 & \multicolumn{3}{c}{Macro F1} & \multicolumn{3}{c}{F1 (G5)} \\
\cmidrule(lr){2-4}\cmidrule(lr){5-7}
Encoder & Single-scale & Fusion & $\Delta$ (abs) & Single-scale & Fusion & $\Delta$ (abs) \\
\midrule
Virchow2 \cite{zimmermann2024virchow2}         & $0.717$ & $\mathbf{0.852}$ & $+0.135$ & $0.514$ & $\mathbf{0.730}$ & $+0.216$ \\
DeiT-S (ViT)~\cite{touvron2021training}        & $0.635$ & $\mathbf{0.799}$ & $+0.164$ & $0.466$ & $\mathbf{0.661}$ & $+0.195$ \\
ResNet-50 (CNN)~\cite{he2016deep}              & $0.547$ & $\mathbf{0.781}$ & $+0.234$ & $0.401$ & $\mathbf{0.592}$ & $+0.191$ \\
FastVim-S (ViM)~\cite{kapse2025fastvim}        & $0.649$ & $\mathbf{0.824}$ & $+0.175$ & $0.524$ & $\mathbf{0.656}$ & $+0.132$ \\
\bottomrule
\end{tabular}%
}
\par\vspace{2pt}
{\footnotesize $\Delta$ (abs): absolute improvement (Fusion $-$ Single-scale).}
\end{table}

\begin{figure}[t]\centering
\includegraphics[width=\textwidth]{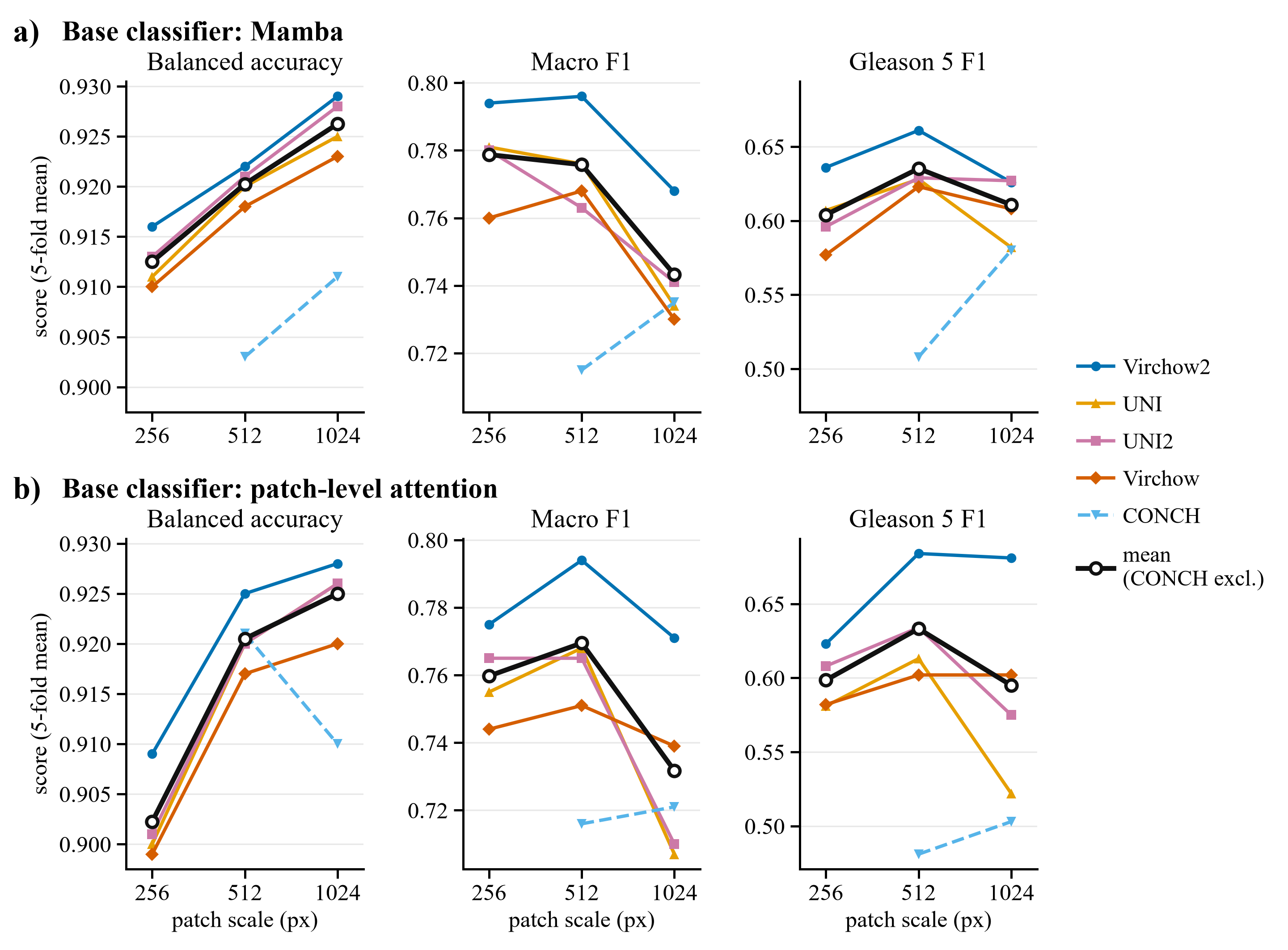}
\caption{Change in evaluation metrics across different scales and foundation models, for two base classifiers. (a) Mamba base classifier. (b) Patch-level attention base classifier. Each line is one foundation model, with the mean across models shown in black excluding CONCH, which lacks a $256$\,px run.}
\label{fig:fmhead}
\end{figure}

\subsection{Effect of Scale, Foundation Model, and Base Classifier}
\label{sec:exp:analysis}
This subsection examines the effects of image scale, the foundation model, and the base classifier on single-scale performance. The 5 base classifiers are a single- and a 3-layer MLP, a Mamba3~\cite{mamba3}, and a patch-level attention encoder, each replacing the Mamba encoder. For Virchow2 with the Mamba base classifier, balanced accuracy increases with scale, from $0.916$ at $256$ to $0.929$ at $1024$, while macro F1 peaks at $512$ ($0.796$) and falls at $1024$ ($0.768$), because pooling over larger regions separates the majority classes but dilutes the fine detail the minority classes rely on. The Mamba base classifier gives the highest balanced accuracy at every scale and the highest macro F1 at $256$ and $512$\,px, which is where the minority classes are best separated, so we adopt it in the fused model. The advantage of the Mamba base classifier over a single MLP shrinks from $0.048$ balanced accuracy at $256$ to $0.012$ at $1024$, as larger-scale features already carry regional context. Table~\ref{tab:app:grid} in the supplementary material reports balanced accuracy, macro F1, and Gleason 5 F1 across all FM-classifier-scale combinations.

\FloatBarrier
\section{Conclusion}
\label{sec:conclusion}
We present a computational method for patch-level Gleason grading that encodes FM features at $256$, $512$, and $1024$ scales with Mamba encoders and fuses their representations in one jointly trained network, extended by a distilled FastVim branch that reads the raw image crops. On the PANDA dataset, our method substantially improves macro F1 and minority-class performance over the best single-scale model while keeping balanced accuracy competitive.

Our method has two practical advantages. First, the same fusion improves macro F1 and Gleason 5 F1 for foundation model, vision transformer, convolutional and ViM encoders (Table~\ref{tab:agnostic}), so it is agnostic to the underlying encoder. Second, current pathology FMs typically operate on a fixed input size, preventing direct encoding of multiple image scales. Our FastVim branch instead processes the native image crops at each scale and improves over the FM branch on every metric, while running faster than a vision transformer of the same size (Table~\ref{tab:backbone}) and requiring far less training data and computation than building an FM.

The main limitation is a slightly lower balanced accuracy than the best single-scale, as our model trades some majority-class recall for large gains on the minority (Gleason 5) class. Upon inspection, part of this drop comes from boundary patches that are mislabelled as cancer regions (Sec.~\ref{sec:exp:main}), so we prioritise macro F1 and minority-class F1, which better reflect performance on the rare high-grade tissue. Our method can be further extended in future work to include validation across additional cohorts and scanners, and to develop a self-supervised pathology state-space feature extractor that eliminates the mean pooling step.
\begin{credits}
\subsubsection{\ackname}
This work is supported by the Agency for Science, Technology and Research (A*STAR), Singapore with Grant ID H25J6a0011 under the Industrial Alignment Fund Pre-positioning Programme (AIF-PP). The computational work is performed on resources provided by the National Supercomputing Centre, Singapore.

\subsubsection{\discintname}
The authors have no competing interests to declare that are relevant to the content of this article.
\end{credits}

\bibliographystyle{splncs04}
\bibliography{references}

\newpage
\appendix
\setcounter{table}{0}
\setcounter{figure}{0}
\renewcommand{\thetable}{S\arabic{table}}
\renewcommand{\thefigure}{S\arabic{figure}}
\section*{Supplementary Material}

\begin{figure}[!h]\centering
\includegraphics[width=\textwidth]{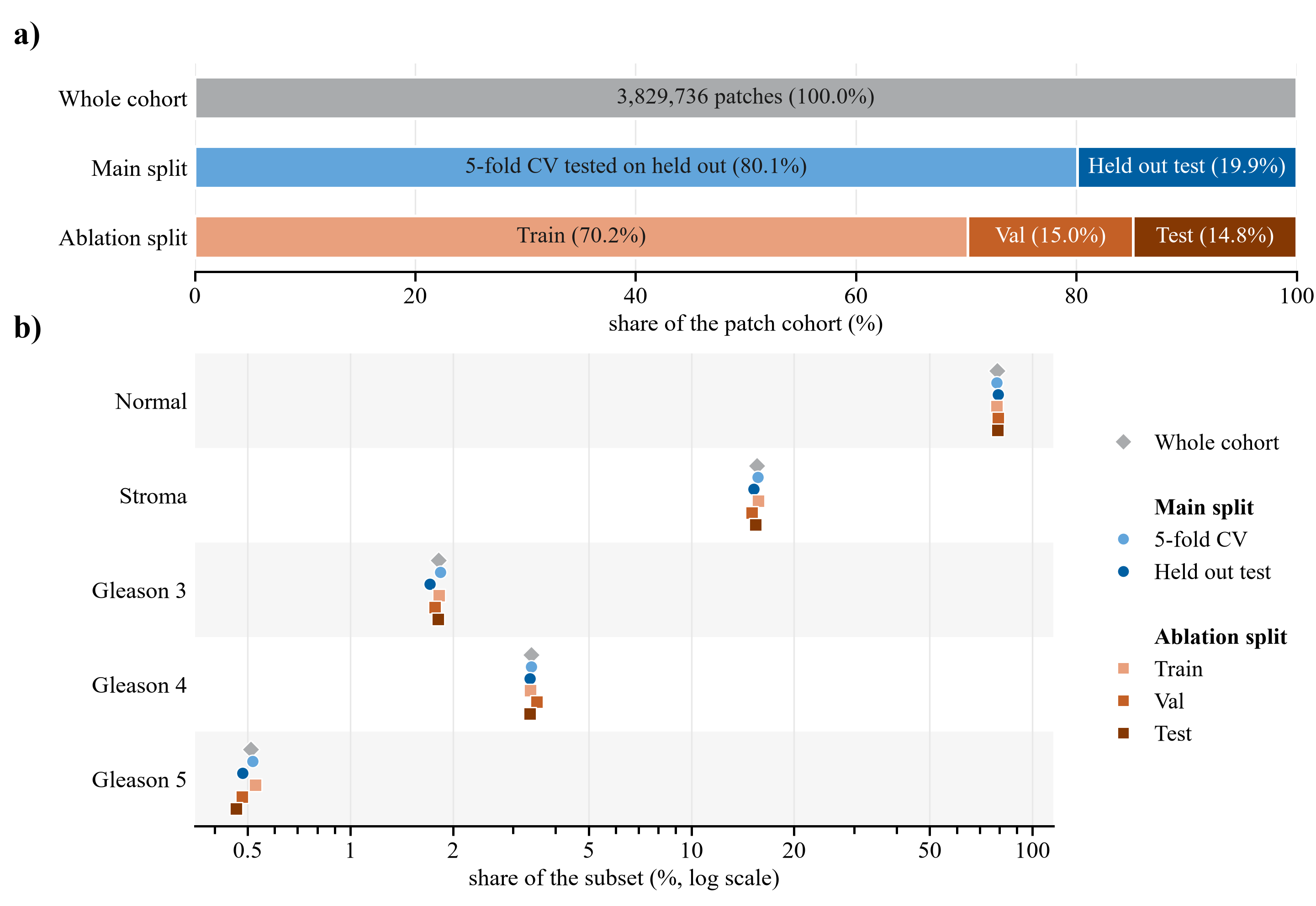}
\caption{Data splits and their class distribution. (a) The main split, used for the 5-fold cross-validation experiments presented in Sec.~\ref{sec:exp:main} and tested on held-out test split. The ablation split, the single $70/15/15$ train/validation/test split used in Sec.~\ref{sec:exp:analysis}. (b) Class distribution within each split, relative to the whole cohort, on a log scale.}
\label{fig:splits}
\end{figure}

\begin{table}[!h]
\centering
\caption{Evaluation metrics across foundation models, base classifiers, and input scales. Results report the mean held-out test performance across 5-fold cross-validation experiments. BA is balanced accuracy, MF1 is macro F1, and G5 is the F1 of Gleason\,5. Since CONCH requires an input size of at least $512$ px, results are reported only for patch sizes of $512$ px and above. No multi-scale fusion is applied in these experiments.}
\label{tab:app:grid}
\setlength{\tabcolsep}{2.5pt}
\scriptsize
\begin{tabular}{llccc ccc ccc}
\toprule
& & \multicolumn{3}{c}{$256$\,px} & \multicolumn{3}{c}{$512$\,px} & \multicolumn{3}{c}{$1024$\,px} \\
\cmidrule(lr){3-5}\cmidrule(lr){6-8}\cmidrule(lr){9-11}
FM & Base classifier & BA & MF1 & G5 & BA & MF1 & G5 & BA & MF1 & G5 \\
\midrule
\multirow{5}{*}{Virchow2}
 & Single-layer MLP      & 0.869 & 0.753 & 0.622 & 0.897 & 0.754 & 0.613 & 0.917 & 0.767 & 0.658 \\
 & 3-layer MLP           & 0.872 & 0.741 & 0.600 & 0.898 & 0.760 & 0.646 & 0.915 & \B{0.776} & \B{0.683} \\
 & Mamba                 & \B{0.916} & \B{0.794} & \B{0.636} & 0.922 & \B{0.796} & 0.661 & \B{0.929} & 0.768 & 0.626 \\
 & Mamba3                & 0.912 & 0.772 & 0.572 & 0.923 & 0.780 & 0.636 & 0.928 & 0.725 & 0.565 \\
 & Patch-level attention & 0.909 & 0.775 & 0.623 & \B{0.925} & 0.794 & \B{0.684} & 0.928 & 0.771 & 0.681 \\
\midrule
\multirow{5}{*}{UNI}
 & Single-layer MLP      & 0.853 & 0.719 & 0.546 & 0.887 & 0.747 & 0.605 & 0.905 & \B{0.763} & 0.638 \\
 & 3-layer MLP           & 0.857 & 0.708 & 0.544 & 0.891 & 0.731 & 0.588 & 0.910 & 0.756 & \B{0.651} \\
 & Mamba                 & 0.911 & \B{0.781} & \B{0.607} & \B{0.920} & 0.776 & 0.628 & 0.925 & 0.734 & 0.582 \\
 & Mamba3                & \B{0.913} & 0.774 & 0.597 & 0.917 & \B{0.777} & \B{0.641} & \B{0.928} & 0.744 & 0.585 \\
 & Patch-level attention & 0.900 & 0.755 & 0.581 & \B{0.920} & 0.768 & 0.613 & 0.926 & 0.707 & 0.522 \\
\midrule
\multirow{5}{*}{UNI2}
 & Single-layer MLP      & 0.863 & 0.729 & 0.578 & 0.893 & 0.736 & 0.586 & 0.911 & \B{0.758} & \B{0.655} \\
 & 3-layer MLP           & 0.866 & 0.721 & 0.554 & 0.895 & 0.738 & 0.609 & 0.914 & 0.757 & 0.626 \\
 & Mamba                 & \B{0.913} & \B{0.780} & 0.596 & \B{0.921} & 0.763 & 0.629 & \B{0.928} & 0.741 & 0.627 \\
 & Mamba3                & 0.911 & 0.779 & \B{0.616} & 0.917 & \B{0.772} & 0.622 & 0.924 & 0.750 & 0.636 \\
 & Patch-level attention & 0.901 & 0.765 & 0.608 & 0.920 & 0.765 & \B{0.634} & 0.926 & 0.710 & 0.575 \\
\midrule
\multirow{5}{*}{Virchow}
 & Single-layer MLP      & 0.855 & 0.727 & 0.573 & 0.885 & 0.729 & 0.579 & 0.907 & 0.721 & 0.557 \\
 & 3-layer MLP           & 0.861 & 0.723 & 0.574 & 0.889 & 0.743 & 0.620 & 0.911 & 0.750 & 0.616 \\
 & Mamba                 & \B{0.910} & 0.760 & 0.577 & \B{0.918} & \B{0.768} & \B{0.623} & \B{0.923} & 0.730 & 0.608 \\
 & Mamba3                & 0.906 & \B{0.770} & \B{0.586} & 0.915 & 0.764 & 0.617 & 0.919 & \B{0.776} & \B{0.661} \\
 & Patch-level attention & 0.899 & 0.744 & 0.582 & 0.917 & 0.751 & 0.602 & 0.920 & 0.739 & 0.602 \\
\midrule
\multirow{5}{*}{CONCH}
 & Single-layer MLP      & --- & --- & --- & 0.884 & 0.711 & 0.530 & 0.904 & 0.723 & 0.561 \\
 & 3-layer MLP           & --- & --- & --- & 0.874 & 0.707 & \B{0.540} & 0.897 & \B{0.738} & 0.576 \\
 & Mamba                 & --- & --- & --- & 0.903 & 0.715 & 0.508 & \B{0.911} & 0.735 & \B{0.580} \\
 & Mamba3                & --- & --- & --- & 0.903 & \B{0.717} & 0.536 & 0.904 & 0.733 & 0.577 \\
 & Patch-level attention & --- & --- & --- & \B{0.921} & 0.716 & 0.481 & 0.910 & 0.721 & 0.503 \\
\bottomrule
\end{tabular}
\end{table}

\end{document}